\documentclass[sigconf]{acmart}

\AtBeginDocument{%
  }

\usepackage{bbold}  
\usepackage{pifont}
\usepackage{bm}
\usepackage{multirow}
\usepackage{colortbl} 

\definecolor{purple2}{HTML}{5523BE}
\definecolor{blue2}{HTML}{0F8FE6}
\definecolor{red2}{HTML}{D64781}

\newcommand{\mytilde}{\raisebox{0.5ex}{\texttildelow}}
\newcommand{\tss}[1]{\scalebox{0.8}{\texttt{#1}}}

\copyrightyear{2025}
\acmYear{2025}
\setcopyright{cc}
\setcctype{by}
\acmConference[MM '25]{Proceedings of the 33rd ACM International Conference on Multimedia}{October 27--31, 2025}{Dublin, Ireland}
\acmBooktitle{Proceedings of the 33rd ACM International Conference on Multimedia (MM '25), October 27--31, 2025, Dublin, Ireland}
\acmDOI{10.1145/3746027.3754566}
\acmISBN{979-8-4007-2035-2/2025/10}

\acmSubmissionID{mfp7307}

\begin{document}

\settopmatter{printacmref=false} 
\setcopyright{none}             
\renewcommand\footnotetextcopyrightpermission[1]{}  
\pagestyle{plain}   
\fancyhead{}        

\title{Vector-Quantized Vision Foundation Models for Object-Centric Learning}

\author{Rongzhen Zhao}
\authornote{Corresponding author.}
\orcid{0009-0000-3964-7336}
\affiliation{
\department{Department of Electrical Engineering and Automation}
\institution{Aalto University}
\city{Espoo}
\country{Finland}
}
\email{rongzhen.zhao@aalto.fi}

\author{Vivienne Huiling Wang}
\orcid{0000-0002-3871-1858}
\affiliation{
\department{Department of Electrical Engineering and Automation}
\institution{Aalto University}
\city{Espoo}
\country{Finland}
}
\email{vivienne.wang@aalto.fi}

\author{Juho Kannala}
\orcid{0000-0001-5088-4041}
\affiliation{
\department{Department of Computer Science}
\institution{Aalto University}
\city{Espoo}
\country{Finland}
}
\affiliation{
\department{Center for Machine Vision and Signal Analysis}
\institution{University of Oulu}
\city{Oulu}
\country{Finland}
}
\email{juho.kannala@aalto.fi}

\author{Joni Pajarinen}
\orcid{0000-0003-4469-8191}
\affiliation{
\department{Department of Electrical Engineering and Automation}
\institution{Aalto University}
\city{Espoo}
\country{Finland}
}
\email{joni.pajarinen@aalto.fi}

\renewcommand{\shortauthors}{Rongzhen Zhao, Vivienne Huiling Wang, Juho Kannala, and Joni Pajarinen}

\begin{abstract}
Object-Centric Learning (OCL) aggregates image or video feature maps into object-level feature vectors, termed \textit{slots}. It's self-supervision of reconstructing the input from slots struggles with complex object textures, thus Vision Foundation Model (VFM) representations are used as the aggregation input and reconstruction target.
Existing methods leverage VFM representations in diverse ways yet fail to fully exploit their potential.
In response, we propose a unified architecture, Vector-Quantized VFMs for OCL (VQ-VFM-OCL, or VVO).
The key to our unification is simply shared quantizing VFM representations in OCL aggregation and decoding.
Experiments show that across different VFMs, aggregators and decoders, our VVO consistently outperforms baselines in object discovery and recognition, as well as downstream visual prediction and reasoning.
We also mathematically analyze why VFM representations facilitate OCL aggregation and why their shared quantization as reconstruction targets strengthens OCL supervision.
Our source code and model checkpoints are available on https://github.com/Genera1Z/VQ-VFM-OCL.
\end{abstract}



\maketitle

\begin{figure}[]
\raggedright
\includegraphics[width=0.99\linewidth]{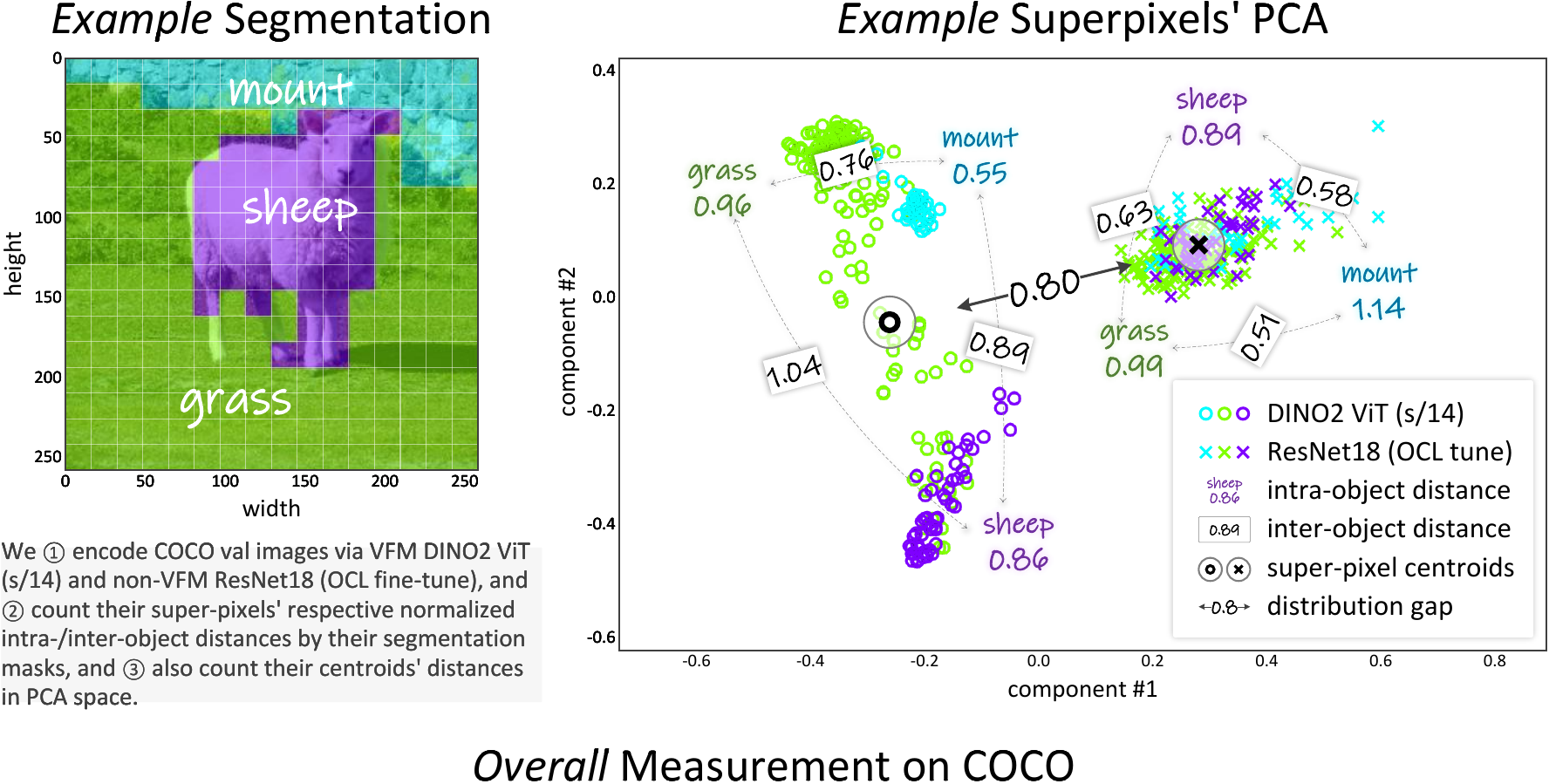}
\vspace{-0.65\baselineskip}
\footnotesize\sffamily
\setlength{\tabcolsep}{1.75pt}
\setlength{\aboverulesep}{0pt}  
\setlength{\belowrulesep}{0pt}  

\begin{tabular}{cccccc}
\hline
     & \multicolumn{2}{c}{DINO2 ViT (s/14)} & \multicolumn{2}{c}{ResNet18 (OCL fine-tune)} & distribution shift \\
\cmidrule(lr){2-3}   \cmidrule(lr){4-5}
     & intra-obj dist & inter-obj dist & intra-obj dist & inter-obj dist & (\textit{in PCA space}) \\
\cmidrule(){1-6}
COCO & 0.6810 & 0.8772 & 0.8028 & 0.7984 & 0.7358 \\
\hline
\end{tabular}

\caption{\textmd{
We utilize two observations:
(\textit{i}) Compared to non-VFMs, VFMs extract features with better object separability, i.e., smaller intra-object distances and larger inter-object distances $\Rightarrow$ We facilitate OCL aggregation via VFM features;
(\textit{ii}) VFM and non-VFM features have a distribution gap, i.e., separated centroids $\Rightarrow$ 
We strengthen OCL supervision by reconstructing the quantized features shared from the same VFM rather than from another encoder.
}}
\label{fig:teaser}
\end{figure}

\section{Introduction}
\label{sect:introduction}

Objects can form highly diverse visual scenes through arrangements and combinations. But mainstream methods based on feature patches or a single feature vector disregard such compositionality. Inspired by human vision cognition \cite{bar2004visual, cavanagh2011visual, palmeri2004visual}, Object-Centric Learning (OCL) decomposes visual scenes into multiple feature vectors, known as \textit{slots}, each corresponding to an object or the background, thus enabling improved modeling of relationships and dynamics among objects. Object-centric representations have demonstrated superiority in advanced vision tasks, like prediction, reasoning, planning, and decision-making \cite{wu2022slotformer}, as well as in interactions between visual modality and other modalities \cite{zhang2024omgllava, wang2024omnidrive}.

Existing OCL methods typically adopt an encoder-aggregator-decoder architecture \cite{locatello2020slotattent}. Firstly, the encoder transforms input image or video frame pixels into a dense feature map. Then, the aggregator sparsifies this feature map into feature vectors via Slot Attention \cite{locatello2020slotattent} with initial slots as the query. Lastly, the decoder reconstructs the input in some form from these aggregated slots, to provide the self-supervised training signal.

OCL relies on pixel textures to discover objects.
The early milestone \cite{locatello2020slotattent} reconstructs input pixels as supervision, usually failing on realistic objects. Some \cite{kipf2021savi, elsayed2022savipp} reconstruct optical flow or depth map to mitigate textural noises, at the cost of expensive annotations. Some \cite{singh2021slate, singh2022steve} reconstruct input's VAE (Variational Autoencoder) representation, whose super-pixels are codebook codes, thus suppressing pixel redundancy and facilitating aggregation from features into slots. 
Recent advances \cite{seitzer2023dinosaur, wu2023slotdiffuz} use Vision Foundation Models (VFMs) \cite{caron2021dino1, oquab2023dino2} to extract features with better object separability, boosting OCL significantly.

However, existing OCL methods leverage VFM representations in quite different ways, as shown in Figure~\ref{fig:compare_model}, and none of them fully utilize the power of VFM representations.

\begin{figure*}[]
\centering
\includegraphics[width=0.975\linewidth]{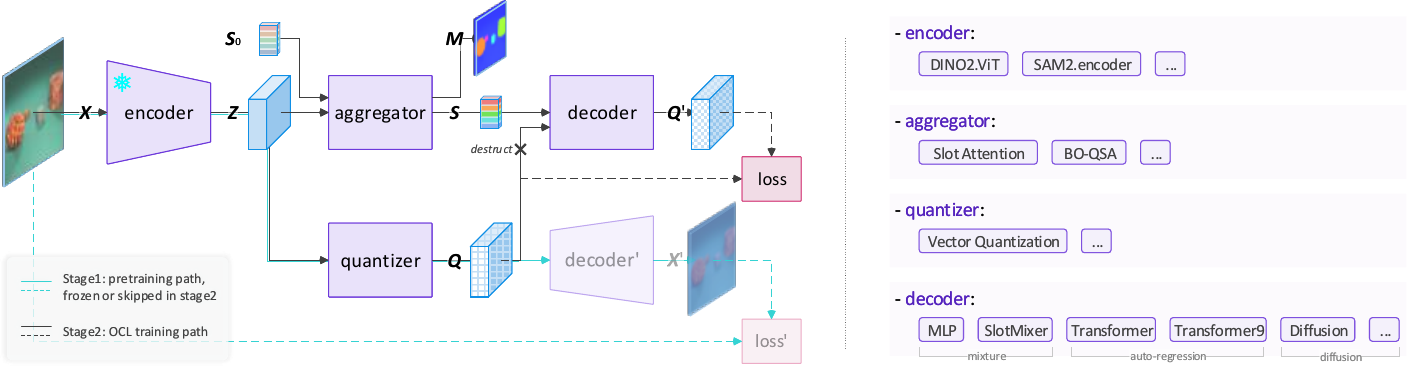}
\caption{\textmd{
VVO is a unified architecture that fully utilizes VFMs in OCL. It not only extracts VFM features with better objectness to facilitate object information aggregation; but further quantizes those VFM features as reconstruction targets to strengthen OCL training supervision. With typical SlotAttention or it variants as the \textcolor{purple2}{aggregator} and Vector-Quantization as the \textcolor{purple2}{quantizer}, VVO supports different VFMs as the \textcolor{purple2}{encoder}, and supports mainstream mixture, auto-regression and diffusion models as the \textcolor{purple2}{decoder}.
}}
\label{fig:vvo_arch}
\end{figure*}

To address these issues, we propose a clean architecture--Vector-Quantized VFMs for OCL (VQ-VFM-OCL, or VVO)--that unifies mainstream OCL methods.
In this architecture, as shown in Figure~\ref{fig:vvo_arch}, our VVO supports different VFMs for encoding, different OCL aggregators and different OCL decoders.
The key to such unification is very simple--We quantize the representations from the same VFM, rather than from another encoder, as the reconstruction target.
We also mathematically analyze why VFM representations facilitate OCL aggregation and why their shared quantization as reconstruction targets strengthens OCL supervision.

Our contributions are: 
(\textit{i}) A clean architecture, which unifies mainstream OCL methods.
(\textit{ii}) Shared quantized VFM representations as reconstruction targets, which not only supports various OCL decoders but also boosts performance.
(\textit{iii}) Insights of why VFM features facilitate OCL aggregation and their shared quantization as reconstruction targets strengthens OCL supervision.

\section{Related Work}
\label{sect:related_work}


\textit{OCL encoding}. 
Early milestone methods like IODINE\cite{greff2019iodine} and SA \cite{locatello2020slotattent} use small naive CNNs \cite{krizhevsky2012alexnet} as OCL encoder. 
Followups like SAVi \cite{kipf2021savi}, SAVi++ \cite{elsayed2022savipp}, SLATE \cite{singh2021slate} and STEVE \cite{singh2022steve} employ pretrained ResNets \cite{he2016resnet}, and fine-tune them on OCL datasets. 
State-of-the-arts like SlotDiffusion \cite{wu2023slotdiffuz} and DINOSAUR \cite{seitzer2023dinosaur} utilize VFMs like DINO \cite{caron2021dino1} and DINO2 \cite{oquab2023dino2} ViTs (Vision Transformers) to extract highly object-separable feature map from input pixels, improving OCL performance significantly.
SAM \cite{kirillov2023sam} and SAM2 \cite{ravi2024sam2} are also well recognized VFMs yet remain unexploited in OCL setting. 
Our VVO supports various VFMs for OCL encoding.

\textit{OCL aggregation}. SlotAttention \cite{locatello2020slotattent} is the footstone for mainstream OCL methods. Subsequent works like BO-QSA \cite{jia2023boqsa}, ISA \cite{biza2023isa} and SysBind \cite{singh2022sysbind} are all its variants, which are designed without changing the external interface. But considering their performance boosts, we only integrate BO-QSA by default.

\textit{OCL decoding}. 
With SlotAttention as the aggregator, the decoder and its reconstruction target affect OCL performance the most, as it is the source of supervision.
Mixture-based decoding, used in SAVi, SAVi++, DINOSAUR and VideoSAUR \cite{zadaianchuk2024videosaur}, decodes each slot's spatial broadcast \cite{watters2019spatialbroadcast} using naive CNNs or MLPs, and mixes them with corresponding weights into the reconstruction.
Transformer-based decoding, used in SLATE, STEVE and SPOT \cite{kakogeorgiou2024spot}, reconstructs VAE representation of the input auto-regressively with slots as the condition.
Diffusion-based decoding in LSD \cite{jiang2023lsd} and SlotDiffusion drives slots to recover noise added to the input's VAE representation.
Our VVO supports all these types of OCL decoding.

\textit{VAE for OCL}. 
Variational Autoencoders (VAEs), like dVAE \cite{im2017dvae} in SLATE and VQ-VAE \cite{van2017vqvae} in SlotDiffusion, are employed to produce reconstruction targets for OCL training.
Since these VAEs are designed for image generation, some methods adapt them for OCL.
Inspired by channel or weight grouping, GDR \cite{zhao2024gdr} decomposes features into attributes and combine them to produce VAE representation as reconstruction targets to guide OCL better.
MSF \cite{zhao2024msf} firstly exploits the multi-scale idea in the OCL setting with VAE-specific designs.
Based on recent advancement RVQ \cite{yang2023rvq} and SimVQ \cite{zhu2024simvq}, we design our own VQ variant for OCL.

\section{Proposed Method}
\label{sect:proposed_method}

We propose Vector-Quantized Vision Foundation Models for Object-Centric Learning, or VQ-VFM-OCL (VVO), elegantly unifying mainstream OCL and consistently boosting their performance.

\subsection{Unify OCL}
\label{sect:background}

Our method adopts an architectural design as shown in Figure~\ref{fig:vvo_arch}.

\textit{Firstly}, OCL \textcolor{purple2}{encoder} $\bm{\phi}_{\mathrm{e}}$ transforms \textcolor{blue2}{input}, an image or video frame $\bm{X} \in \mathbb{R} ^ {h_0 \times w_0 \times c_0}$ of some visual scene, into a dense \textcolor{blue2}{feature map} $\bm{Z} \in \mathbb{R} ^ {h \times w \times c}$, for the following query-based aggregation:
\begin{equation}
\label{eq:ocl_encode}
\bm{\phi}_{\mathrm{e}} : \bm{X} \rightarrow \bm{Z}
\end{equation}
where $\bm{\phi}_{\mathrm{e}}$ can be parameterized as pretrained VFMs, like DINO \cite{caron2021dino1}, DINO2 \cite{oquab2023dino2}, SAM \cite{kirillov2023sam} and SAM2 \cite{ravi2024sam2}. As OCL relies on textures to separate objects, $\bm{\phi}_{\mathrm{e}}$ should handle complex textures of objects, making VFMs necessary here. We will explain it in Section~\ref{sect:maximize_vfms_in_ocl}.

\textit{Secondly}, given \textcolor{blue2}{queries} $\bm{S}_0 \in \mathbb{R} ^ {n \times c}$, OCL \textcolor{purple2}{aggregator} $\bm{\phi}_{\mathrm{a}}$ transforms $\bm{Z}$ into multiple feature vectors or \textcolor{blue2}{slots} $\bm{S} \in \mathbb{R} ^ {n \times c}$ and corresponding byproduct \textcolor{blue2}{segmentation} masks $\bm{M} \in \mathbb{R} ^ {h \times w}$, each representing a specific object or background in the scene:
\begin{equation}
\label{eq:ocl_aggregat}
\bm{\phi}_{\mathrm{a}} : (\bm{S}_0, \bm{Z}) \rightarrow (\bm{S}, \bm{M})
\end{equation}
where $\bm{\phi}_{\mathrm{a}}$ can be parameterized as widely adopted SlotAttention \cite{locatello2020slotattent} and its variants, which is some cross attention with $\bm{S}_0$ as queries and $\bm{Z}$ as keys and values. $\bm{M}$ is the binarized attention map thus intuitively reflects how well objects are represented by slots. 

For video OCL, there is a recurrent module transitioning current slots $\bm{S}$ into new queries $\bm{S}_0$ for the next time step. Such module can be parameterized as a Transformer encoder block \cite{vaswani2017transformer}.

\textit{Meanwhile}, OCL \textcolor{purple2}{quantizer} $\bm{\phi}_{\mathrm{q}}$ transforms $\bm{X}$ into the reconstruction \textcolor{blue2}{target} $\bm{Q} \in \mathbb{R} ^ {h \times w \times c}$ for the following decoding:
\begin{equation}
\label{eq:ocl_quantiz}
\bm{\phi}_{\mathrm{q}} : \bm{X} \rightarrow \bm{Q}
\end{equation}
where $\bm{\phi}_{\mathrm{q}}$ can be parameterized as some Vector Quantization (VQ) \cite{im2017dvae, van2017vqvae}. But we meticulously design our own VQ variant, as detailed in Section~\ref{sect:maximize_vfms_in_ocl}.
$\bm{\phi}_{\mathrm{q}}$ is pretrained in VAE framework and is frozen afterwards, where the encoder is shared from the frozen OCL encoder $\bm{\phi}_{\mathrm{e}}$, and the decoder is a typical VAE decoder. We will explain why not use a separate typical VAE encoder in Section~\ref{sect:maximize_vfms_in_ocl}.

\textit{Thirdly}, OCL \textcolor{purple2}{decoder} $\bm{\phi}_{\mathrm{d}}$ transforms slots $\bm{S}$ into \textcolor{blue2}{reconstruction} $\bm{Q}' \in \mathbb{R} ^ {h \times w \times c}$ with destructed $\bm{Q}$ as the condition:
\begin{equation}
\label{eq:ocl_decode}
\bm{\phi}_{\mathrm{d}} : (\bm{S}, \mathrm{destruct}(\bm{Q}) ) \rightarrow \bm{Q}'
\end{equation}
Here $\bm{\phi}_{\mathrm{d}}$ can be parameterized as 
(\textit{i}) a CNN or MLP for mixture decoding \cite{kipf2021savi, seitzer2023dinosaur}, where $\bm{Q}$ is destructed to height and width, and $\bm{S}$ is spatially broadcast \cite{watters2019spatialbroadcast} into this shape and then decoded into components being mixed together; 
(\textit{ii}) or a Transformer decoder for auto-regressive decoding \cite{singh2021slate, kakogeorgiou2024spot}, where $\bm{Q}$ is destructed with causal masking as the query and $\bm{S}$ is the key and value; 
(\textit{iii}) or a conditional Diffusion model for diffusion decoding \cite{wu2023slotdiffuz, jiang2023lsd}, where $\bm{Q}$ is destructed with noise as the input and $\bm{S}$ is the condition.

Reconstructing $\bm{Q}$ using $\bm{S}$ drives $\bm{S}$ to aggregate as much object information as possible. Thus, a good reconstruction target, like in \cite{zhao2024gdr, zhao2024msf}, is very important. We will explain this in Section~\ref{sect:maximize_vfms_in_ocl}.

\textit{Lastly}, the supervision signal for OCL training comes from minimizing the reconstruction \textcolor{red2}{loss} between $\bm{Q}'$ and $\bm{Q}$:
\begin{equation}
\label{eq:ocl_loss}
\mathrm{min} _ {(\bm{\phi}_{\mathrm{a}}, \bm{\phi}_{\mathrm{d}})}  f_{\mathrm{recon}} (\bm{Q}', \bm{Q})
\end{equation}
where $f_{\mathrm{recon}}(\cdot,\cdot)$ can be (\textit{i}) Mean Squared Error (MSE) loss for mixture and diffusion OCL decoding, or (\textit{ii}) Cross Entropy (CE) loss for auto-regressive decoding.

\subsection{Utilize VFMs in OCL}
\label{sect:maximize_vfms_in_ocl}

In our unified architecture, we utilize VFMs as following.

\textbf{\textit{Direct VFM Feature Extraction for Better Aggregation}}

We directly extract the feature map $\bm{Z}$ from the input $\bm{X}$ using VFMs as $\bm{\phi}_{\mathrm{e}}$, where DINO2 ViT and SAM2 encoder are chosen and experimented thoroughly. No extra position encoding is needed here like in SA \cite{locatello2020slotattent} because these VFMs already contain the positional information required in $\bm{\phi}_{\mathrm{a}}$. Since $\bm{\phi}_{\mathrm{e}}$ is frozen, We further use a trainable linear layer to adjust $\bm{Z}$ slightly. 

As shown in Figure~\ref{fig:teaser}, VFM representations have better objectness than non-VFMs, even under naive kMeans clustering\footnote{https://scikit-learn.org/stable/modules/generated/sklearn.cluster.KMeans.html}. 
OCL aggregation is essentially a clustering whose initial centroids are trainable \cite{jia2023boqsa}. 
Thus, we expect $\bm{\phi}_{\mathrm{a}}$ to aggregate VFM's $\bm{Z}$ into slots $\bm{S}$ better under queries $\bm{S}_0$.
Previous methods like DINOSAUR \cite{seitzer2023dinosaur} have already exploited this but without any reason.

\textbf{\textit{Shared VFM Feature Quantization for Better Supervision}}

Given the VFM feature $\bm{Z}$, we adjust it via a small CNN to eliminate the positional information for better quantization; Then quantize it using our VQ variant $\bm{\phi}_{\mathrm{q}}$ as the reconstruction target $\bm{Q}$.

Our VQ's codebook follows SimVQ \cite{zhu2024simvq}. We predefine $m=4096$ template features, i.e., a codebook $\bm{T}_0 \in \mathbb{R} ^ {m \times c_0}$, which are randomly initialized and remain frozen.
In vector quantization, we project $\bm{T}$ with a pre-trainable linear layer and match it with the adjusted $\bm{Z}$:
\begin{equation}
\label{eq:codebook_project}
\bm{T} := \bm{W} \cdot \mathrm{sg}(\bm{T}_0)
\end{equation}
\begin{equation}
\label{eq:codebook_match}
\bm{D} = || \bm{Z} - \bm{T} || _ 2 ^ 2
\end{equation}
where $\mathrm{sg}(\cdot)$ is stop-gradient; $\bm{T} \in \mathbb{R} ^ {m \times c}$ is the codebook for quantizing $\bm{Z}$; $\bm{D} \in \mathbb{R} ^ {h \times w \times m}$ is the matching distance between every super-pixel in $\bm{Z}$ and every code in $\bm{T}$.

We convert distances to probabilities and select the most matched codes to form the quantization $\bm{Q}$ as the reconstruction target:
\begin{equation}
\label{eq:softmax}
\bm{P} = \mathrm{softmax}_c ( -\bm{D} )
\end{equation}
\begin{equation}
\label{eq:argmax}
\bm{I} = \mathrm{argmax}_m (\bm{P})
\end{equation}
\begin{equation}
\label{eq:select}
\bm{Q} = \mathrm{index}_m (\bm{T}, \bm{I})
\end{equation}
where $\mathrm{softmax}(\cdot)$ is calculated along the channel dimension; $\bm{P}$ is the match probabilities; $\bm{I} \in \mathbb{R} ^ {h \times w}$ is the matched code indexes; $\mathrm{argmax}(\cdot)$ is calculated along the channel dimension; $\mathrm{index}(\cdot, \cdot)$ is operated along code number dimension.
The typical STE \cite{bengio2013ste} on $\bm{Q}$, needed in pre-training, can be skipped during OCL training.

For its pre-training, we introduce some tricks. 
We add noise to $\bm{D}$ before Equation~\ref{eq:softmax} to encourage code utilization:
\begin{equation}
\label{eq:gumbel}
\bm{D} := \frac{\bm{D}+\bm{G}}{\tau}
\end{equation}
where $\bm{G} \in \mathbb{R}^{h \times w \times m}$ is Gumbel noise and $\tau$ is the temperature.
Training-time annealing residual connection \cite{zhao2024gdr, zhao2024msf} is added after Equation~\ref{eq:select} to stabilize pre-training:
\begin{equation}
\label{eq:residual}
\bm{Q} := \alpha \bm{Z} + (1-\alpha) \bm{Q}
\end{equation}
where $\alpha$ is scheduled from 1 to 0 during pre-training using cosine-annealing.
Besides typical losses of reconstruction, alignment and commitment \cite{van2017vqvae}, we regularize the adjusted $\bm{Z}$ to be normal:
\begin{equation}
\label{eq:regulariz}
l_{\mathrm{n}} = \lambda \mathrm{MSE} ( \bm{Z}, \mathrm{sg}( \frac { \bm{Z} - \mathbb{E}[\bm{Z}] } { \sqrt{ \mathbb{V}[\bm{Z}] + \epsilon } } ) )
\end{equation}
where $\lambda$ is empirically set to 0.1; $\mathbb{E}$ and $\mathbb{V}$ are calculated along height, width and channel dimensions.

With all samples' feature maps $\bm{Z}$ being represented with one codebook $\bm{T}$, the quantization $\bm{Q}$ naturally gains cross-sample consistency, helping the aggregation with queries $\bm{S}_0$, which are also shared across samples. 
Such tokenization is compatible with both regression and classification decoding. 
In contrast, methods like SLATE \cite{singh2021slate} and SlotDiffusion \cite{wu2023slotdiffuz} are faced with distribution gaps between $\bm{Q}$ and $\bm{Z}$, shown in Figure~\ref{fig:teaser}, due to separate VAE and OCL encoders.
Thus, we expect shared VFM representation quantization as reconstruction targets to strengthen OCL supervision.

\begin{figure}[]
\centering
\includegraphics[width=\linewidth]{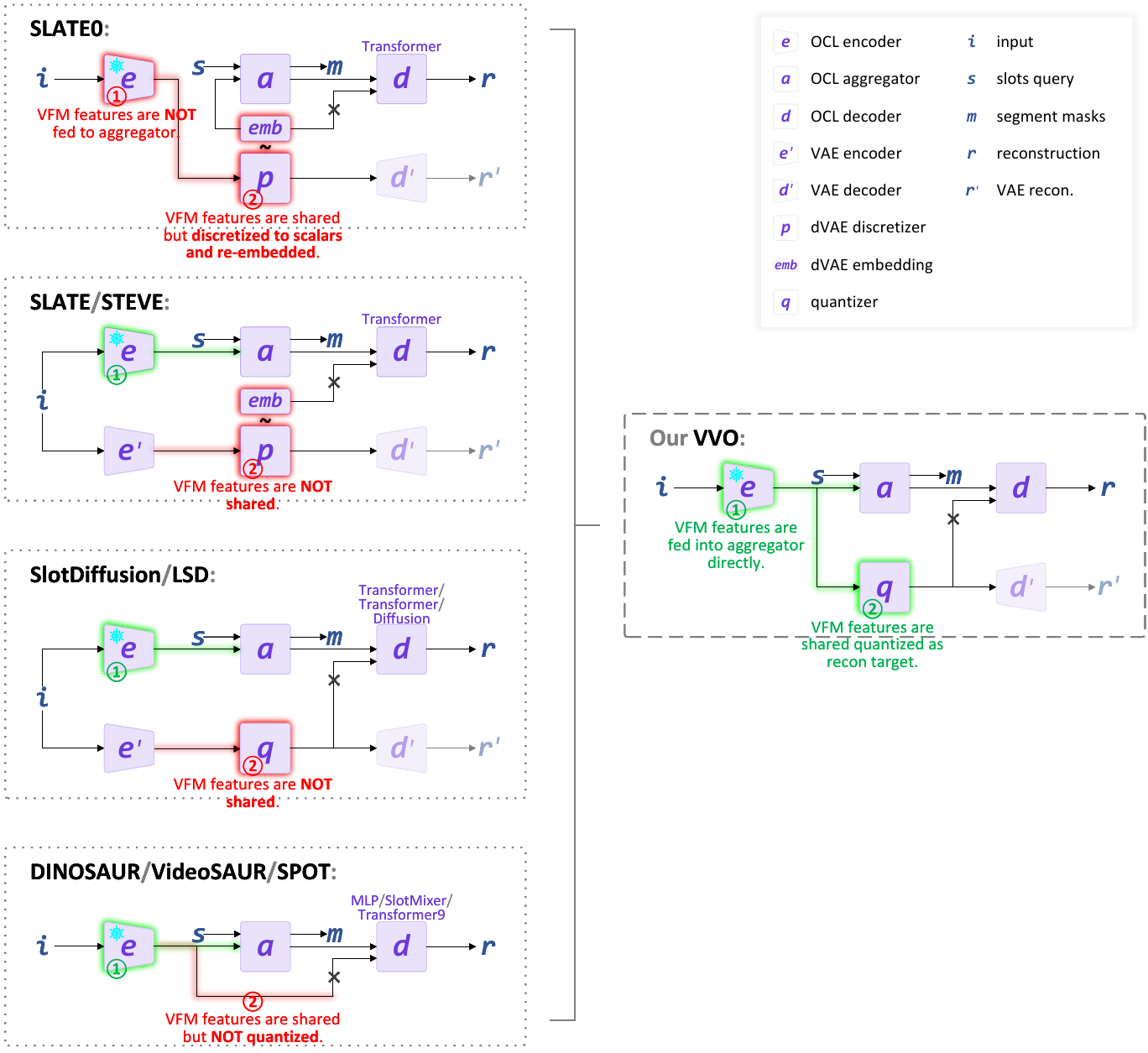}
\caption{\textmd{
Model architecture comparison.
}}
\label{fig:compare_model}
\end{figure}

\begin{figure*}[]
\centering
\includegraphics[width=0.99\linewidth]{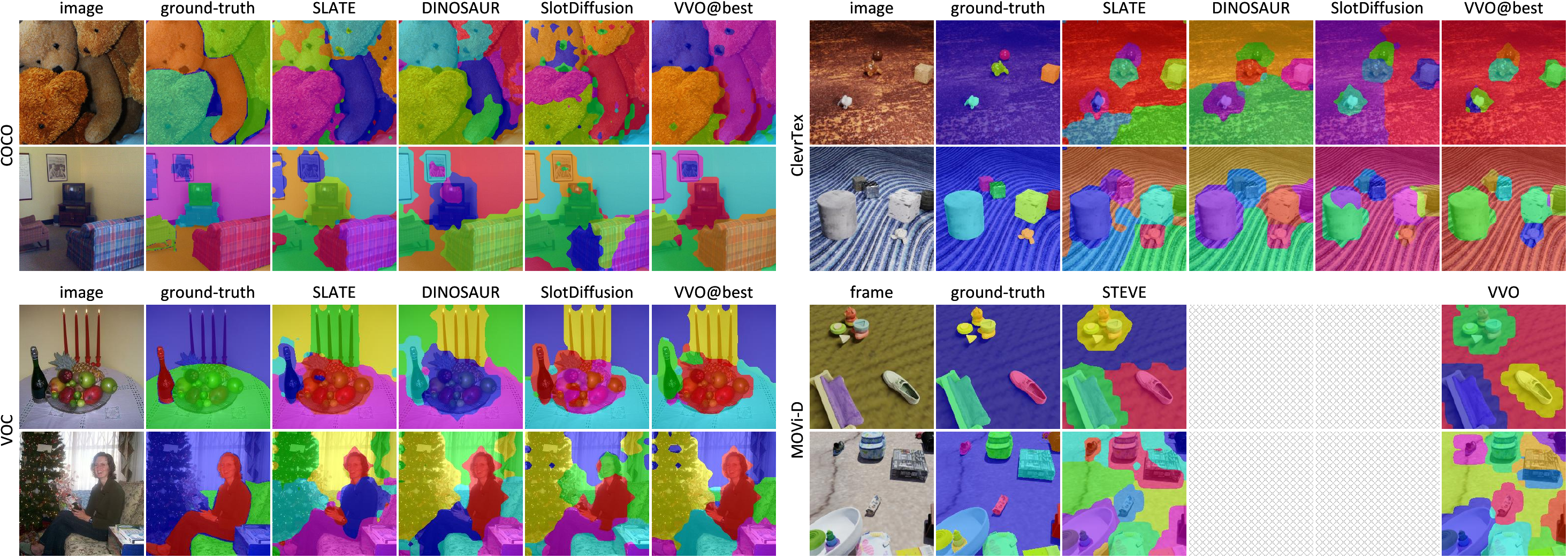}
\caption{\textmd{
Qualitative object discovery performance comparison. 
}}
\label{fig:qualitative}
\end{figure*}

\begin{table*}[]
\centering
\small
\setlength{\tabcolsep}{1.5pt}
\setlength{\aboverulesep}{0pt}  
\setlength{\belowrulesep}{0pt}  
\newcommand{\cg}[1]{\textcolor{green}{#1}}
\newcommand{\std}[1]{\scalebox{0.4}{±#1}}

\begin{tabular}{ccccccccccccccccc}
\hline
\textcolor{gray}{resolution=} & \multicolumn{4}{c}{ClevrTex {\tiny \#slot=11}} & \multicolumn{4}{c}{COCO {\tiny \#slot=7}} & \multicolumn{4}{c}{VOC {\tiny \#slot=6}} & \multicolumn{4}{c}{MOVi-D {\tiny \#slot=21 conditional}} \\
\arrayrulecolor{gray}
\cmidrule(lr){2-5} \cmidrule(lr){6-9} \cmidrule(lr){10-13} \cmidrule(lr){14-17}
\arrayrulecolor{black}
\textcolor{gray}{256$\times$256 (224)} & ARI & ARI\textsubscript{fg} & mBO & mIoU & ARI & ARI\textsubscript{fg} & mBO & mIoU & ARI & ARI\textsubscript{fg} & mBO & mIoU & ARI & ARI\textsubscript{fg} & mBO & mIoU \\
\cmidrule(){1-17}
SLATE-DINO      & 17.4\std{2.9}  & 87.4\std{1.7} & 44.5\std{2.2} & 43.3\std{2.4} & 17.5\std{0.6} & 28.8\std{0.3} & 26.8\std{0.3} & 25.4\std{0.3} & 18.6\std{0.1} & 26.2\std{0.8} & 37.2\std{0.5} & 36.1\std{0.4} & -- & -- & -- & -- \\
VQDINO\tss{Tfd} & \cg{55.4}\std{18.2} & 85.9\std{0.7} & \cg{54.4}\std{2.4} & \cg{53.6}\std{2.5} & \cg{21.1}\std{2.1} & \cg{31.5}\std{1.1} & \cg{29.6}\std{0.7} & \cg{28.2}\std{0.8} & \cg{22.7}\std{0.9} & \cg{26.6}\std{1.3} & \cg{39.8}\std{0.3} & \cg{38.9}\std{0.3} & -- & -- & -- & -- \\
\arrayrulecolor{gray}
\cmidrule(lr){1-1} \cmidrule(lr){2-5} \cmidrule(lr){6-9} \cmidrule(lr){10-13} \cmidrule(lr){14-17}
STEVE-DINO      & -- & -- & -- & -- & -- & -- & -- & -- & -- & -- & -- & -- & 33.7\std{0.2} & 66.2\std{0.3} & 22.6\std{0.2} & 20.9\std{0.3} \\
VQDINO\tss{TfdT}    & -- & -- & -- & -- & -- & -- & -- & -- & -- & -- & -- & -- & \cg{37.2}\std{0.7} & \cg{72.7}\std{4.9} & \cg{25.7}\std{1.5} & \cg{24.3}\std{1.7} \\
\cmidrule(lr){1-1} \cmidrule(lr){2-5} \cmidrule(lr){6-9} \cmidrule(lr){10-13} \cmidrule(lr){14-17}
DINOSAUR-DINO   & 50.7\std{24.1} & 89.4\std{0.3} & 53.3\std{5.0} & 52.8\std{5.2} & 18.2\std{1.0} & 37.0\std{1.2} & 28.3\std{0.5} & 26.9\std{0.5} & 21.5\std{0.7} & 36.2\std{1.3} & 40.6\std{0.6} & 39.7\std{0.6} & -- & -- & -- & -- \\
VQDINO\tss{Mlp}    & \cg{66.0}\std{0.3}  & 88.7\std{0.6} & \cg{57.0}\std{0.5} & \cg{56.6}\std{0.5} & \cg{19.2}\std{0.4} & 35.9\std{0.6} & \cg{28.7}\std{0.3} & \cg{27.4}\std{0.3} & \cg{21.7}\std{0.8} & 35.4\std{1.0} & \cg{41.0}\std{0.1} & \cg{40.1}\std{0.2} & -- & -- & -- & -- \\
\cmidrule(lr){1-1} \cmidrule(lr){2-5} \cmidrule(lr){6-9} \cmidrule(lr){10-13} \cmidrule(lr){14-17}
SlotDiffusion-DINO & 66.1\std{1.3}  & 82.7\std{1.6} & 54.3\std{0.5} & 53.4\std{0.8} & 17.7\std{0.5} & 29.0\std{0.1} & 27.0\std{0.4} & 25.6\std{0.4} & 17.0\std{1.2} & 21.7\std{1.8} & 35.2\std{0.9} & 34.0\std{1.0} & -- & -- & -- & -- \\
VQDINO\tss{Dfz} & \cg{72.2}\std{0.2}  & 81.9\std{2.0} & \cg{57.6}\std{0.7} & \cg{56.8}\std{0.8} & \cg{18.3}\std{0.4} & 28.7\std{1.0} & \cg{27.2}\std{0.1} & \cg{25.8}\std{0.1} & \cg{19.1}\std{0.5} & \cg{24.2}\std{0.3} & \cg{36.9}\std{0.1} & \cg{35.9}\std{0.2} & -- & -- & -- & -- \\
\arrayrulecolor{black}
\hline
\end{tabular}

\caption{\textmd{
Object discovery performance with \textbf{DINO2 ViT (s/14)} for OCL encoding. VVO is instantiated as VQDINO; \texttt{Tfd}, \texttt{TfdT}, \texttt{Mlp} and \texttt{Dfz} are Transformer, Transformer-temporal, MLP and Diffusion for OCL decoding respectively. 
}}
\label{tab:quantitative_vqdino}
\end{table*}

\begin{table*}[]
\centering
\small
\setlength{\tabcolsep}{1.5625pt}
\setlength{\aboverulesep}{0pt}  
\setlength{\belowrulesep}{0pt}  
\newcommand{\cg}[1]{\textcolor{green}{#1}}
\newcommand{\std}[1]{\scalebox{0.4}{±#1}}

\begin{tabular}{ccccccccccccccccc}
\hline
\textcolor{gray}{resolution=} & \multicolumn{4}{c}{ClevrTex {\tiny \#slot=11}} & \multicolumn{4}{c}{COCO {\tiny \#slots=7}} & \multicolumn{4}{c}{VOC {\tiny \#slot=6}} & \multicolumn{4}{c}{MOVi-D {\tiny \#slot=21 conditional}} \\
\arrayrulecolor{gray}
\cmidrule(lr){2-5} \cmidrule(lr){6-9} \cmidrule(lr){10-13} \cmidrule(lr){14-17}
\arrayrulecolor{black}
\textcolor{gray}{256$\times$256} & ARI & ARI\textsubscript{fg} & mBO & mIoU & ARI & ARI\textsubscript{fg} & mBO & mIoU & ARI & ARI\textsubscript{fg} & mBO & mIoU & ARI & ARI\textsubscript{fg} & mBO & mIoU \\
\cmidrule(){1-17}
SLATE-SAM       & 18.0\std{0.3} & 82.8\std{0.9} & 51.8\std{0.8} & 50.6\std{0.8} & 12.5\std{0.8} & 27.4\std{1.4} & 22.7\std{0.8} & 21.5\std{0.8} & 16.3\std{0.7} & 21.4\std{1.3} & 34.8\std{0.5} & 33.9\std{0.6} & -- & -- & -- & -- \\
VQSAM\tss{Tfd}  & \cg{19.9}\std{4.1} & \cg{87.3}\std{1.1} & \cg{53.3}\std{2.4} & \cg{51.6}\std{2.8} & \cg{16.4}\std{0.7} & 26.3\std{0.5} & \cg{26.6}\std{0.5} & \cg{25.2}\std{0.6} & \cg{18.6}\std{1.0} & \cg{22.2}\std{1.8} & \cg{37.2}\std{0.6} & \cg{36.0}\std{0.6} & -- & -- & -- & -- \\
\arrayrulecolor{gray}
\cmidrule(lr){1-1} \cmidrule(lr){2-5} \cmidrule(lr){6-9} \cmidrule(lr){10-13} \cmidrule(lr){14-17}
STEVE-SAM       & -- & -- & -- & -- & -- & -- & -- & -- & -- & -- & -- & -- & 31.2\std{3.8} & 62.9\std{5.1} & 22.2\std{2.2} & 19.9\std{2.4} \\
VQSAM\tss{TfdT} & -- & -- & -- & -- & -- & -- & -- & -- & -- & -- & -- & -- & \cg{31.3}\std{3.7} & \cg{63.8}\std{2.5} & \cg{22.6}\std{1.3} & \cg{20.5}\std{1.4} \\
\cmidrule(lr){1-1} \cmidrule(lr){2-5} \cmidrule(lr){6-9} \cmidrule(lr){10-13} \cmidrule(lr){14-17}
DINOSAUR-SAM    & 21.1\std{0.8} & 88.7\std{0.6} & 46.3\std{0.5} & 45.3\std{0.5} & 11.7\std{2.6} & 21.6\std{1.4} & 21.8\std{2.5} & 20.6\std{2.5} & 12.1\std{0.4} & 15.6\std{1.8} & 29.4\std{0.6} & 28.7\std{0.6} & -- & -- & -- & -- \\
VQSAM\tss{Mlp}  & \cg{64.1}\std{1.1} & 87.4\std{0.7} & \cg{58.8}\std{0.7} & \cg{58.0}\std{0.8} & \cg{13.2}\std{0.3} & \cg{25.2}\std{1.2} & \cg{23.5}\std{0.4} & \cg{22.2}\std{0.3} & \cg{14.2}\std{0.1} & \cg{19.4}\std{0.6} & \cg{32.6}\std{0.2} & \cg{31.6}\std{0.3} & -- & -- & -- & -- \\
\cmidrule(lr){1-1} \cmidrule(lr){2-5} \cmidrule(lr){6-9} \cmidrule(lr){10-13} \cmidrule(lr){14-17}
SlotDiffusion-SAM & 19.5\std{0.5} & 81.2\std{0.6} & 55.1\std{0.4} & 53.6\std{0.4} & 16.8\std{0.5} & 27.1\std{0.7} & 26.7\std{0.4} & 25.3\std{0.4} & 17.0\std{0.9} & 18.9\std{2.1} & 36.2\std{0.6} & 34.9\std{0.8} & -- & -- & -- & -- \\
VQSAM\tss{Dfz}  & \cg{30.0}\std{0.3} & 78.6\std{0.8} & \cg{57.3}\std{0.2} & \cg{56.1}\std{0.2} & \cg{17.3}\std{0.4} & 26.0\std{0.2} & \cg{27.2}\std{0.4} & \cg{25.8}\std{0.4} & \cg{17.3}\std{0.5} & \cg{20.2}\std{2.3} & \cg{36.3}\std{0.8} & \cg{35.1}\std{0.9} & -- & -- & -- & -- \\
\arrayrulecolor{black}
\hline
\end{tabular}

\caption{\textmd{
Object discovery performance with \textbf{SAM2 Hiera+FPN (t/16)} for OCL decoding. VVO is instantiated as VQSAM; \texttt{Tfd}, \texttt{TfdT}, \texttt{Mlp} and \texttt{Dfz} are Transformer, Transformer-temporal, MLP and Diffusion for OCL decoding respectively. 
}}
\label{tab:quantitative_vqsam}
\end{table*}

\subsection{Compare Architectures}
\label{sect:compare_models}

As shown in Figure~\ref{fig:compare_model}, we compare baselines methods with our VVO in a unified perspective.
Specifically,

- \textit{Our VVO}: (1) VFMs are employed for OCL encoding and their features are fed to the aggregator directly, which eases OCL aggregation, as formulated in Section~\ref{sect:maximize_vfms_in_ocl}. (2) VFM features are shared quantized as reconstruction targets, which strengthens OCL self supervision, as formulated in Section~\ref{sect:maximize_vfms_in_ocl}.

- \textit{SLATE0} (the official version) \cite{singh2021slate}: (1) VFM features are \textbf{NOT directly fed to} the aggregator. (2) VFM features are shared \textbf{discretized} to scalar numbers and re-embedded into features to be learned latter. This loses much information of VFM features.

- \textit{SLATE} (the improved version; adopted here) / \textit{STEVE} \cite{jia2023boqsa,singh2022steve,wu2023slotdiffuz}: (1) Same as VVO. (2) Reconstruction targets are \textbf{discretized} from features of \textbf{separate} VAE encoding, not quantized from features shared from OCL encoding, causing optimization noises.

- \textit{SlotDiffusion} \cite{wu2023slotdiffuz} / \textit{LSD} \cite{jiang2023lsd}: (1) Same as VVO. (2) Reconstruction targets are quantized from \textbf{separate} VAE encoding features, not sharing OCL encoding features, causing optimization noises.

- \textit{DINOSAUR} \cite{seitzer2023dinosaur} / \textit{VideoSAUR} \cite{zadaianchuk2024videosaur} / \textit{SPOT} \cite{kakogeorgiou2024spot}: (1) Same as VVO. (2) Reconstruction targets are shared from VFM features of OCL encoding \textbf{without quantization}, causing optimization noises.

Our VVO realizes a clean and unified architecture based on the above-mentioned two key designs.
Specifically, 

- \textit{Mixture-based OCL decoders}, e.g., CNN \cite{locatello2020slotattent,kipf2021savi,elsayed2022savipp}, MLP \cite{seitzer2023dinosaur} and SlotMixer \cite{zadaianchuk2024videosaur}, are originally designed for continuous features, thus are compatible with our shared quantized VFM features. 

- \textit{Auto-regressive OCL decoders}, e.g., the Transformer decoder \cite{singh2021slate,singh2022steve} and Transformer9 \cite{kakogeorgiou2024spot}, are designed for discretized features, while also showing applicability to continuous features \cite{seitzer2023dinosaur,kakogeorgiou2024spot}. Thus they are applicable to our shared quantized VFM features.

- \textit{Diffusion-based OCl decoders}, e.g., conditional Diffusion \cite{wu2023slotdiffuz,jiang2023lsd}, work on low-dimension features, necessitating our shared quantization on the continuous high-dimensional VFMs features.

\section{Experiment}
\label{sect:experiment}

We conduct all experiments using three random seeds.

\subsection{Set up the Benchmark}

\textit{Datasets}. 
We include both synthetic and real-world datasets.
ClevrTex\footnote{https://www.robots.ox.ac.uk/{\mytilde}vgg/data/clevrtex} comprises synthetic images, each with about 10 geometric objects scattered in complex background.
MOVi-D\footnote{https://github.com/google-research/kubric/blob/main/challenges/movi/README.md\#\\movi-d} contains synthetic videos, each with up to 20 daily objects dropping and bumping.
COCO\footnote{https://cocodataset.org} is a recognized real-world image dataset, and we use its instance segmentation.
VOC\footnote{http://host.robots.ox.ac.uk/pascal/VOC} is a real-world image dataset, and we use its instance segmentation.
We also report results on real-world video dataset YTVIS\footnote{https://youtube-vos.org/dataset/vis} version HQ\footnote{https://github.com/SysCV/vmt?tab=readme-ov-file\#hq-ytvis-high-quality-video-instance-segmentation-dataset}, which contains large-scale short videos from YouTube.
We choose Physion\footnote{https://physion-benchmark.github.io} for visual prediction and reasoning as it contains common object interactions, requiring algorithms to learn dynamics like support, roll and link, then to predict and reason about future scene states.

\textit{Models}. 
We compare VVO with both OCL classics and state-of-the-arts.
SLATE \cite{singh2021slate} uses a Transformer decoder for auto-regressive decoding, and it \textcolor{gray}{differs} from VVO in a separate VAE encoder and naive quantizer; STEVE \cite{singh2022steve} is SLATE's video version.
DINOSAUR \cite{seitzer2023dinosaur} uses an MLP for mixture decoding, and it \textcolor{gray}{differs} from VVO in no quantization in its reconstruction target.
SlotDiffusion \cite{wu2023slotdiffuz} uses a conditional Diffusion model for diffusion decoding, and it \textcolor{gray}{differs} from VVO in a separate VAE encoder and naive quantizer.
General improvers GDR \cite{zhao2024gdr} and MSF \cite{zhao2024msf} only support auto-regression and diffusion decoding.
We skip outdated methods like IODINE \cite{greff2019iodine}, SA \cite{locatello2020slotattent} and ISA \cite{biza2023isa} due to their low accuracy. We also skip SAVi \cite{kipf2021savi} and SAVi++ \cite{elsayed2022savipp} as their extra modalities are unfair to others.

\textit{Comparison}. Instead of copying existing results, we reproduce all baselines to realize fair comparison. 
We use identical data augmentation, VFMs in OCL encoding and training recipes for all experiment items unless not applied.
We instantiate all baselines' VAE part as TAESD\footnote{https://huggingface.co/docs/diffusers/en/api/models/autoencoder\_tiny}, which is a large-scale pretrained StableDiffusion\footnote{https://huggingface.co/spaces/stabilityai/stable-diffusion} module, to build all \textit{strong} baselines.

\begin{table}[]
\centering
\small
\setlength{\tabcolsep}{4pt}
\setlength{\aboverulesep}{0pt}  
\setlength{\belowrulesep}{0pt}  
\newcommand{\cg}[1]{\textcolor{green}{#1}}
\newcommand{\std}[1]{\scalebox{0.4}{±#1}}

\begin{tabular}{ccccc}
\hline
 & ARI & ARI\textsubscript{fg} & mBO & mIoU \\
\arrayrulecolor{gray}
\cmidrule(lr){2-5}
\arrayrulecolor{black}


\multicolumn{5}{c}{Using \textbf{higher resolution}: 384$\times$384 (336)}\\
\hline
\textcolor{gray}{resolution=384$\times$384 (336)} & \multicolumn{4}{c}{\textcolor{gray}{COCO \tiny\#slot=7}}\\
\arrayrulecolor{gray}
\cmidrule(lr){1-1} \cmidrule(lr){2-5}
SLATE-DINO         & 41.4\std{1.0} & 34.0\std{0.3} & 27.4\std{0.4} & 25.9\std{0.5} \\
VQDINO\tss{Tfd}         & \cg{44.1}\std{0.8} & \cg{37.5}\std{1.1} & \cg{29.6}\std{0.5} & \cg{28.0}\std{0.5} \\
\cmidrule(lr){1-1} \cmidrule(lr){2-5}
DINOSAUR-DINO      & 45.0\std{0.1} & 42.2\std{0.5} & 29.9\std{0.1} & 28.5\std{0.1} \\
VQDINO\tss{Mlp}         & 44.6\std{0.7} & \cg{42.6}\std{0.5} & 29.8\std{0.3} & \cg{28.6}\std{0.3} \\
\cmidrule(lr){1-1} \cmidrule(lr){2-5}
SlotDiffusion-DINO & 41.6\std{0.5} & 34.5\std{0.4} & 27.7\std{0.2} & 26.2\std{0.2} \\
VQDINO\tss{Dfz}         & \cg{43.4}\std{1.3} & 34.2\std{0.4} & \cg{28.3}\std{0.7} & \cg{26.9}\std{0.7} \\
\arrayrulecolor{black}
\hline


\multicolumn{5}{c}{Using \textbf{different aggregators}: SlotAttention, BO-QSA}\\
\hline
\textcolor{gray}{resolution=256$\times$256 (224)} & \multicolumn{4}{c}{\textcolor{gray}{COCO \tiny\#slot=7}}\\
\arrayrulecolor{gray}
\cmidrule(lr){1-1} \cmidrule(lr){2-5}
\textcolor{gray}{SLATE-DINO}-SlotAttention      & 17.0\std{1.3} & 28.3\std{0.5} & 26.4\std{0.4} & 25.1\std{0.3} \\
\textcolor{gray}{VQDINO\tss{Tfd}}-SlotAttention & \cg{20.8}\std{2.0} & \cg{31.5}\std{1.2} & \cg{29.4}\std{0.9} & \cg{27.9}\std{1.1} \\
\cmidrule(lr){1-1} \cmidrule(lr){2-5}
\textcolor{gray}{SLATE-DINO}-BO-QSA              & 17.5\std{0.6} & 28.8\std{0.3} & 26.8\std{0.3} & 25.4\std{0.3} \\
\textcolor{gray}{VQDINO\tss{Tfd}}-BO-QSA         & \cg{21.1}\std{2.1} & \cg{31.5}\std{1.1} & \cg{29.6}\std{0.7} & \cg{28.2}\std{0.8} \\
\arrayrulecolor{black}
\hline
\end{tabular}

\caption{\textmd{
VVO using higher resolution (\textit{upper}) and different aggregators (\textit{lower}) on object discovery.
By default, we use BO-QSA for all our experiment items, including the baselines.
}}
\label{tab:vqdino_using}
\end{table}

\begin{table}[]
\centering
\small
\setlength{\tabcolsep}{4pt}
\setlength{\aboverulesep}{0pt}  
\setlength{\belowrulesep}{0pt}  
\newcommand{\cg}[1]{\textcolor{green}{#1}}
\newcommand{\std}[1]{\scalebox{0.4}{±#1}}

\begin{tabular}{ccccc}
\hline
 & ARI & ARI\textsubscript{fg} & mBO & mIoU \\
\arrayrulecolor{gray}
\cmidrule(lr){2-5}
\arrayrulecolor{black}


\multicolumn{5}{c}{Compared with \textbf{general improvers}: GDR and MSF}\\
\hline
\textcolor{gray}{resolution=256$\times$256 (224)} & \multicolumn{4}{c}{\textcolor{gray}{COCO \tiny\#slot=7}} \\
\arrayrulecolor{gray}
\cmidrule(lr){1-1} \cmidrule(lr){2-5}
GDR\tss{Tfd}-DINO & 18.0\std{1.4} & 29.2\std{0.2} & 27.4\std{0.7} & 26.0\std{0.7} \\
MSF\tss{Tfd}-DINO & 18.0\std{0.5} & 29.0\std{0.2} & 27.4\std{0.3} & 26.1\std{0.3} \\
VQDINO\tss{Tfd} & \cg{21.1}\std{2.1} & \cg{31.5}\std{1.1} & \cg{29.6}\std{0.7} & \cg{28.2}\std{0.8} \\
\cmidrule(lr){1-1} \cmidrule(lr){2-5}
GDR\tss{Dfz}-DINO & 17.9\std{0.1} & 29.0\std{0.3} & 27.2\std{0.1} & 25.8\std{0.1} \\
MSF\tss{Dfz}-DINO & 16.9\std{0.4} & 28.7\std{0.1} & 26.6\std{0.2} & 25.2\std{0.2} \\
VQDINO\tss{Dfz} & \cg{18.3}\std{0.4} & \cg{28.7}\std{1.0} & \cg{27.2}\std{0.1} & \cg{25.8}\std{0.1} \\
\arrayrulecolor{black}
\hline


\multicolumn{5}{c}{Compared with \textbf{SotA methods}: SPOT, VideoSAUR}\\
\hline
\textcolor{gray}{resolution=256$\times$256 (224)} & \multicolumn{4}{c}{\textcolor{gray}{COCO \tiny\#slot=7}} \\
\arrayrulecolor{gray}
\cmidrule(lr){1-1} \cmidrule(lr){2-5}
SPOT-DINO & 20.3\std{0.7} & 41.1\std{0.3} & 30.4\std{0.1} & 29.0\std{0.9} \\
VQDINO\tss{Tfd9} & \cg{21.3}\std{0.4} & \cg{42.3}\std{1.0} & \cg{31.4}\std{0.2} & \cg{29.9}\std{0.3} \\
\arrayrulecolor{black}
\cmidrule(lr){1-5}
\arrayrulecolor{gray}
\textcolor{gray}{resolution=256$\times$256 (224)} & \multicolumn{4}{c}{\textcolor{gray}{YTVIS (HQ) \tiny\#slot=7, unconditional}} \\
\cmidrule(lr){1-1} \cmidrule(lr){2-5}
VideoSAUR-DINO & 33.0\std{0.6} & 49.0\std{0.9} & 30.8\std{0.4} & 30.1\std{0.6} \\
VQDINO\tss{SmdT} & \cg{35.7}\std{0.5} & \cg{49.5}\std{0.6} & \cg{32.7}\std{0.2} & \cg{31.6}\std{0.5} \\
\arrayrulecolor{black}
\hline
\end{tabular}

\caption{\textmd{
VVO versus general improvers (\textit{upper}) and SotA methods (\textit{lower}) on object discovery.
SPOT uses Transformer with 9 permutations \texttt{Tfd9} as decoder while VideoSAUR uses SlotMixer \texttt{SmdT}.
}}
\label{tab:vqdino_compared}
\end{table}

\begin{table}[]
\centering
\small
\setlength{\aboverulesep}{0pt}  
\setlength{\belowrulesep}{0pt}  
\newcommand{\std}[1]{\scalebox{0.4}{±#1}}
\newcommand{\cg}[1]{\textcolor{green}{#1}}

\begin{tabular}{ccc}
\hline
 & class labels & bounding boxes \\
\cmidrule(lr){2-3}
 & top1$\uparrow$  & R2$\uparrow$   \\
\cmidrule(lr){1-3}
DINOSAUR + MLP & 0.61\std{0.0} & 0.57\std{0.1} \\
VQDINO\tss{Mlp} + MLP & \cg{0.64}\std{0.1} & \cg{0.59}\std{0.1} \\
\hline
\end{tabular}

\caption{\textmd{
Set prediction performance on COCO (\#slot=7).
}}
\label{tab:set_prediction}
\end{table}

\subsection{Evaluate on Object Discovery}
\label{sect:object_discovery}

Object discovery task intuitively shows how well those slots separate different objects.
We evaluate all methods' byproduct object segmentation accuracy with Adjusted Rand Index\footnote{https://scikit-learn.org/stable/modules/generated/sklearn.metrics.adjusted\_rand\_sco\\re.html} (ARI), ARI\textsubscript{fg} (foreground), mean Intersection-over-Union\footnote{https://scikit-learn.org/stable/modules/generated/sklearn.metrics.jaccard\_score.html} (mIoU) and mean Best Overlap \cite{uijlings2013selectivesearch} (mBO) as metrics.

\textit{With unsupervised pretrained VFMs for OCL encoding}, i.e., DINO2 ViT (version s/14), our VVO is instantiated as VQDINO. As shown in Table~\ref{tab:quantitative_vqdino}, our method consistently improves object discovery performance across all types of OCL decoding.
With a Transformer decoder \texttt{Tfd} for auto-regressive decoding, VVO significantly outperforms SLATE and STEVE across all datasets. 
With a spatial broadcast MLP decoder \texttt{Mlp} for mixture-based decoding, VVO shows a smaller advantage over DINOSAUR but is still effective.
With a conditional Diffusion model \texttt{Dfz} for diffusion-based decoding, VVO surpasses SlotDiffusion on most datasets.

\textit{With supervised pretrained VFMs for OCL encoding}, i.e., SAM2 Hiera+FPN (version t/16), our VVO is instantiated as VQSAM. As shown in Table~\ref{tab:quantitative_vqsam}, VVO boosts all baselines' object discovery performance across all decoding types on all datasets.

As shown in Table~\ref{tab:vqdino_using}, whether using \textit{higher input resolution} or \textit{different aggregators}, VVO maintains its superiority over baselines.
As shown in Table~\ref{tab:vqdino_compared}, VVO outperforms recent OCL \textit{general improvers}, i.e., GDR and MSF.
VVO also surpasses \textit{state-of-the-arts}, i.e., SPOT \cite{kakogeorgiou2024spot} and VideoSAUR \cite{zadaianchuk2024videosaur}, with their special types of decoding.

\subsection{Evaluate on Set Prediction}
\label{sect:set_prediction}

Set prediction task directly shows how much object information those slots grasp.
We use OCL to represent dataset COCO as slots, and use a small MLP to predict the object class label and bounding box corresponding to each slot by following this work \cite{seitzer2023dinosaur}. We measure top1 accuracy\footnote{https://scikit-learn.org/stable/modules/generated/sklearn.metrics.accuracy\_score.ht\\ml} accuracy of the classified class labels while measure R2 score\footnote{https://scikit-learn.org/stable/modules/generated/sklearn.metrics.r2\_score.html} of the regressed bounding box coordinates.

As shown in Table~\ref{tab:set_prediction}, compared with DINOSAUR, our VVO, i.e., VQDINO\tss{Mlp}, obtains both better object classification and better object bounding box regression. 
Thus, our method extracts better slot representations for objects than the baseline.

\subsection{Deploy to the Downstream}
\label{sect:downstream}

\begin{figure}[]
\centering
\includegraphics[width=0.75\linewidth]{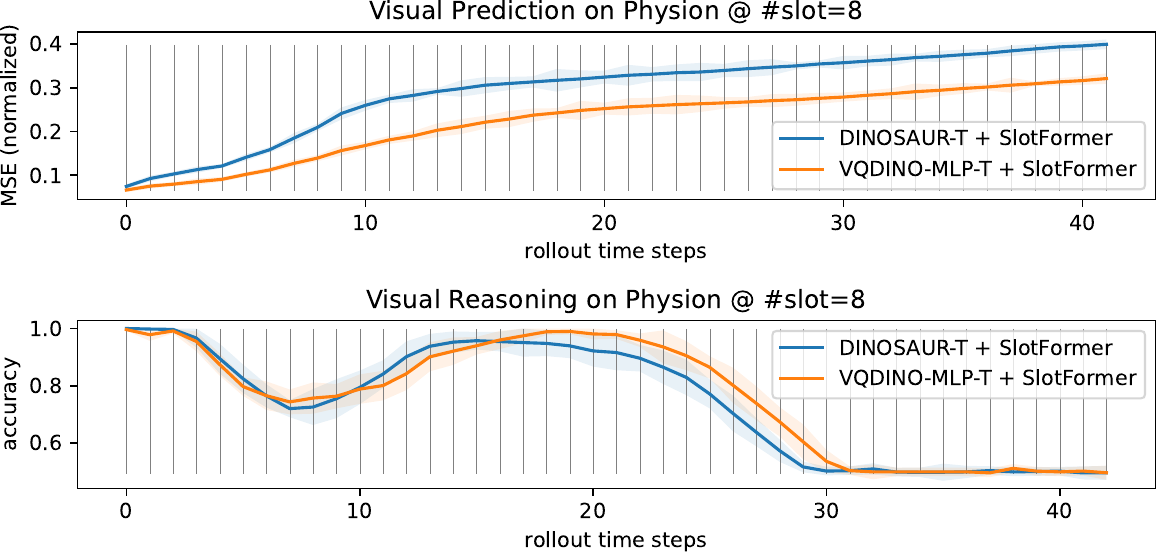}
\caption{\textmd{
Visual prediction (\textit{upper}) and reasoning (\textit{lower}) performance on Physion (\#slot=8).
VVO has smaller prediction error in all time steps, and higher reasoning accuracy in later time steps.
}}
\label{fig:visual_prediction_reasoning}
\end{figure}

Better object representation benefits downstream tasks.
We follow the convention to pretrain OCL models on Physion and represent this dataset as slots. Then the object-centric dynamics model SlotFormer \cite{wu2022slotformer} is trained on those slots in an auto-regressive manner along the time dimension. We use temporal versions of DINOSAUR and our VVO, i.e., VQDINO\tss{Mlp} to extract slots.

On \textit{visual prediction}, we evaluate the per time step prediction errors measured in normalized Mean Squared Error (MSE) between regressed and extracted slots. As shown in Figure~\ref{fig:visual_prediction_reasoning} upper, our VVO accumulates prediction errors slower than the baseline.

On \textit{visual reasoning}, we evaluate the per time step reasoning accuracy between the classification outputs and ground-truth labels.
As shown in Figure~\ref{fig:visual_prediction_reasoning} lower, VVO's accuracies are slightly lower at the beginning but much higher later than the baseline.

\subsection{Ablate the Architecture}

As shown in Table~\ref{tab:ablat_shared_enc}, VVO's design of \textit{shared VAE and OCL encoder} consistently outperforms separate VAE and OCL encoders, even when the latter employs another VFM for VAE encoding. Thus, VVO's design of shared VFM representation quantization is superior to the prevalent design of separate VAE and OCL encoders.

\begin{figure}[]
\centering
\includegraphics[width=0.75\linewidth]{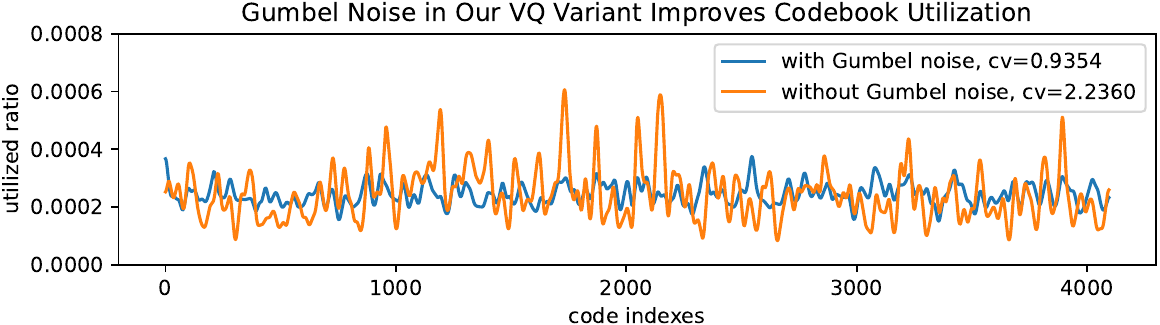}
\\\vspace{0.25\baselineskip}
\includegraphics[width=0.75\linewidth]{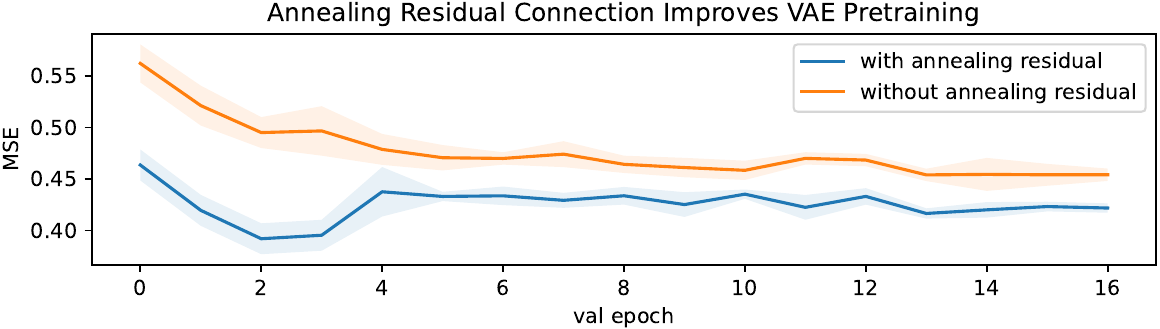}
\caption{\textmd{
Effects of tricks in our VQ variant: (\textit{upper}) Gumbel noise improves codebook utilization, where ``CV'' means Coefficient of Variation, and curves are smoothed by Gaussian kernel of size 10; (\textit{lower}) Annealing residual connection improves VAE pretraining, and the blue curve's turning point at epoch 4 is where the residual connection anneals to zero. Results are from the VAE pretraining of VQDINO\tss{Dfz} on COCO.
}}
\label{fig:ablat_code_utiliz_residual}
\end{figure}

As shown in Table~\ref{tab:ablat_our_vq}, our \textit{improved quantizer variant} for VVO, built upon tricks of Gumbel noises defined in Equation~\ref{eq:gumbel}, annealing residual connection defined in Equation~\ref{eq:residual} and normalizing regularization defined in Equation~\ref{eq:regulariz}, is superior to the naive VQ.
The detailed effects of those tricks are shown in Figure~\ref{fig:ablat_code_utiliz_residual}. \textit{Adding Gumbel noises} increases codebook utilization, contributing to more effective codes; \textit{Annealing residual connection} improves VAE pretraining, contributing to smaller VAE reconstruction error. 

\begin{table}[]
\centering
\small
\setlength{\tabcolsep}{4.5pt}
\setlength{\aboverulesep}{0pt}  
\setlength{\belowrulesep}{0pt}  
\newcommand{\std}[1]{\scalebox{0.4}{±#1}}

\begin{tabular}{ccccc}
\hline
& \multirow{2}{*}{VAE encoding} & \multirow{2}{*}{OCL encoding} & \multicolumn{2}{c}{VQDINO\tss{Tfd}} \\
\arrayrulecolor{gray}
\cmidrule(lr){4-5}
\arrayrulecolor{black}
 &           &                           & ARI   & ARI\textsubscript{fg} \\
\cmidrule(){1-5}
\textit{shared} & DINO2 ViT & DINO2 ViT & 21.1\std{2.1} & 31.5\std{1.1} \\
\arrayrulecolor{gray}
\cmidrule(lr){1-1} \cmidrule(lr){2-3} \cmidrule(lr){4-5}
\multirow{3}{*}{\textit{separate}} & \textcolor{gray}{TAESD encoder} & \textcolor{gray}{ResNet18}  & \textcolor{gray}{15.4\std{3.3}} & \textcolor{gray}{24.1\std{2.5}} \\
\cmidrule(lr){2-3} \cmidrule(lr){4-5}
                & TAESD encoder & DINO2 ViT & 17.5\std{0.6} & 28.8\std{0.3} \\
\cmidrule(lr){2-3} \cmidrule(lr){4-5}
                & SAM2 encoder  & DINO2 ViT & 18.3\std{1.3} & 29.1\std{0.9} \\
\arrayrulecolor{black}
\hline
\end{tabular}

\caption{\textmd{
VVO's two key designs: (\textit{i}) Using VFM representation for encoding is better than using non-VFMs; (\textit{ii}) Sharing the OCL encoder as VAE encoder to obtain targets is better than using separate VAE and OCL encoders.
Results are on COCO.
}}
\label{tab:ablat_shared_enc}
\end{table}

\begin{table}[]
\centering
\small
\setlength{\tabcolsep}{3.5pt}
\setlength{\aboverulesep}{0pt}  
\setlength{\belowrulesep}{0pt}  
\newcommand{\std}[1]{\scalebox{0.4}{±#1}}

\begin{tabular}{cccccc}
\hline
shared VAE / & Gumbel  & annealing & normaliz.   & \multicolumn{2}{c}{VQDINO\tss{Dfz}} \\
\arrayrulecolor{gray}
\cmidrule(lr){5-6}
\arrayrulecolor{black}
OCL encoder & noise   & residual  & regulariz.    & ARI & ARI\textsubscript{fg} \\
\cmidrule(){1-6}
\checkmark & \checkmark & \checkmark & \checkmark & 21.1\std{2.1} & 31.5\std{1.1} \\
\arrayrulecolor{gray}
\cmidrule(lr){1-1} \cmidrule(lr){2-4} \cmidrule(lr){5-6}
\ding{55} & \checkmark & \checkmark & \checkmark & 18.3\std{1.3} & 29.1\std{0.9} \\
\cmidrule(lr){1-1} \cmidrule(lr){2-4} \cmidrule(lr){5-6}
\checkmark & \checkmark &  &  & 19.5\std{0.6} & 29.9\std{1.9} \\
\cmidrule(lr){1-1} \cmidrule(lr){2-4} \cmidrule(lr){5-6}
\checkmark & & \checkmark &  & 20.1\std{1.3} & 29.8\std{1.0} \\
\cmidrule(lr){1-1} \cmidrule(lr){2-4} \cmidrule(lr){5-6}
\checkmark & &  & \checkmark & 20.0\std{0.7} & 29.4\std{0.8} \\
\arrayrulecolor{black}
\hline
\end{tabular}

\caption{\textmd{
VVO's VQ variant: All our three tricks are beneficial to the overall performance boosts. In comparison to VVO's key designs, these tricks are more like the cherry on top.
Results are on COCO with settings consistent with the above.
}}
\label{tab:ablat_our_vq}
\end{table}

\section{Analysis}
\label{sect:analysis}

We mathematically analyze our two key designs as below.

\textbf{Aggregation as Clustering}

\begin{figure}[]
\centering
\includegraphics[width=0.75\linewidth]{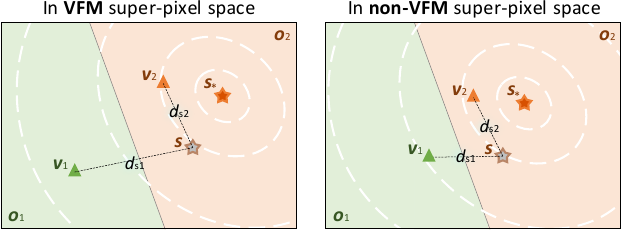}
\caption{\textmd{
Better objectness helps OCL aggregation. Green and orange areas stand for objects $\bm{o}_1$ and $\bm{o}_2$, where $\bm{v}_1$ and $\bm{v}_2$ are arbitrary super-pixels and $\bm{s}_*$ is $\bm{o}_2$'s centroid. But the actual query $\bm{s} \sim \mathcal{N} ( \bm{s}_*, \bm{\sigma}^2 )$. In VFM super-pixel space, the distance $d_{12}$ between $\bm{v}_1$ and $\bm{v}_2$ is larger, i.e., better objectness, thus $\bm{s}$ has higher probability $p(d_{s1}>d_{s2})$ to represent $\bm{o}_2$ correctly, compared with that in non-VFM super-pixel space.
}}
\label{fig:proof1}
\end{figure}

As shown in Figure~\ref{fig:proof1}, super-pixels in a feature map $\bm{Z}$ all belong to two objects $\bm{o}_1$ and $\bm{o}_2$, so two queries are needed for aggregation, which is basically sum of super-pixels weighted by normalized minus distances between queries and super-pixels \cite{locatello2020slotattent}. 

Denote the ideal query of $\bm{o}_2$ as $\bm{s}_*$, which is the clustering center or centroid of $\bm{o}_2$ and is closer to all super-pixels in $\bm{o}_2$ than in $\bm{o}_1$:
\begin{equation}
\label{eq:d_star_12}
d_{*1} = d(\bm{s}_*, \bm{v}_1) > d(\bm{s}_*, \bm{v}_2) = d_{*2}
\end{equation}
where $d(\cdot, \cdot)$ is a distance metric, e.g., minus inner product; $\bm{v}_1$ and $\bm{v}_2$ are arbitrary points in $\bm{o}_1$ and $\bm{o}_2$, respectively.

But the actual query $\bm{s}$ follows $\mathcal{N}( \bm{s}_*, \bm{\sigma}^2\bm{I} )$. Substituting $\bm{s}$ for $\bm{s}_*$, the probability of correct aggregation is:
\begin{equation}
\label{eq:p2}
p_2 = p(d_{s1} > d_{s2}) = \int_{\bm{v} \in \bm{o}_2} \frac{1}{\sqrt{2\pi} \bm{\sigma}} e ^ { -\frac{1}{2} (\frac{\bm{v} - \bm{s}_*} {\bm{\sigma}} ) ^ 2 } d\bm{v}
\end{equation}
where $\bm{o}_2$ always contains $\bm{s}_*$, and is bounded by the separation hyper-plane between $\bm{o}_1$ and $\bm{o}_2$. The closer the boundary is to $\bm{s}_*$ the smaller the value of $p_2$ would be.

According to Figure~\ref{fig:teaser} the first observation, in VFM super-pixel space, points of the same object have smaller distances while points from different objects have larger distances, compared to that in non-VFM space. This means the separation plane is closer to $\bm{s}_*$ in non-VFM space. Thus, $p_2$ in VFM space is bigger.

\textbf{Shared Quantization and Optimization Noise}

We reconstruct $\bm{Q}'$ to approximate the target $\bm{Q}$ ultimately from $\bm{Z}$ via $\bm{\phi}_{\mathrm{a}} \circ \bm{\phi}_{\mathrm{d}}$, denoted as $f$ for simplicity. Under MSE loss, the gradient with respect to $\bm{Z}$ is:
\begin{equation}
\label{eq:vae_grad_wrt_z}
\frac {\partial \mathrm{MSE} (\bm{Q}', \mathrm{sg}(\bm{Q}))} {\partial \bm{Z}} = 2 (\bm{Q}' - \mathrm{sg}(\bm{Q})) \frac {\partial \bm{Q}'} {\partial \bm{Z}}
\end{equation}

We obtain $\bm{Q}$ by quantizing $\bm{Z}$, i.e., $\mathbb{E}[\bm{Z}] = \bm{Q}$, implying that any deviation of $\bm{Q}'=f(\bm{Z})$ from $\bm{Q}$ is due to $f$ and $\bm{Z}$. Assuming $f$ preserves $\bm{Z}$'s statistical properties, we have:
\begin{equation}
\label{eq:expectat_q_prime}
\mathbb{E}[\bm{Q}'] = \mathbb{E}[f(\bm{Z})] \approx \bm{Q}
\end{equation}

Thus the residual error $\bm{Q}' - \mathrm{sg}({\bm{Q}})$ in Equation~\ref{eq:vae_grad_wrt_z} is statistically unbiased and small on average:
\begin{equation}
\label{eq:expectat_q_prime_minus_q}
\mathbb{E}[\bm{Q}' - \mathrm{sg}(\bm{Q})] \approx 0
\end{equation}

But if instead of sharing $\bm{\phi}_{\mathrm{e}}$, we use an extra VAE encoder plus $\bm{\phi}_{\mathrm{q}}$ to obtain the target, denoted as $\bm{Q}_2$, then $\bm{Q}_2 \neq \bm{Q}$ according to Figure~\ref{fig:teaser} the second observation. Substitute $\bm{Q}_2$ into Equation~\ref{eq:vae_grad_wrt_z} and the residual error $\bm{Q}' - \mathrm{sg}(\bm{Q}_2)$ would be systematically biased:
\begin{equation}
\label{eq:expectat_q_primte_minus_q2}
\mathbb{E}[\bm{Q}' - \mathrm{sg}(\bm{Q}_2)] \neq 0
\end{equation}
which increases noise in optimization.

Under CE loss, the gradient with respective to $\bm{Z}$ is:
\begin{equation}
\label{eq:ce_grad_wrt_z}
\frac {\partial \mathrm{CE} (\bm{Q}', \mathrm{sg}(\bm{Q}))} {\partial \bm{Z}} = \frac {\partial f(\bm{Z}) ^ T} {\partial \bm{Z}} (\bm{Q}' - \mathrm{sg}(\bm{Q}))
\end{equation}
where $\frac {\partial f(\bm{Z})} {\partial \bm{Z}}$ is the Jacobian matrix of $f(\bm{Z})$ with respect to $\bm{Z}$. Anyway, this has similar structure to Equation~\ref{eq:vae_grad_wrt_z} and thus does not alter our judgment above.

\section{Conclusion}
\label{sect:conclusion}

We propose a unified architecture VVO for object-centric representation learning. Our VVO supports different well-recognized vision foundation models for OCL encoding and supports mainstream types of OCL decoding. It boosts the existing OCL performance in object discovery significantly, and benefits downstream tasks of visual prediction and reasoning.
VVO has the potential to serve as a general testbed for research related to OCL in the future.

\begin{acks}
We acknowledge the support of Finnish Center for Artificial Intelligence (FCAI), Research Council of Finland flagship program.
We thank the Research Council of Finland for funding the projects ADEREHA (grant no. 353198), BERMUDA (362407) and PROFI7 (352788).
We also appreciate CSC - IT Center for Science, Finland, for granting access to supercomputers Mahti and Puhti, as well as LUMI, owned by the European High Performance Computing Joint Undertaking (EuroHPC JU) and hosted by CSC Finland in collaboration with the LUMI consortium.
Furthermore, we acknowledge the computational resources provided by the Aalto Science-IT project through the Triton cluster.
\end{acks}

\bibliographystyle{ACM-Reference-Format}
\bibliography{acmart}  


\end{document}